\documentclass{article}

     \PassOptionsToPackage{numbers, compress}{natbib}
\usepackage[final]{neurips_2020}

\usepackage[utf8]{inputenc} %
\usepackage[T1]{fontenc}    %

\usepackage{amsmath,amsfonts,bm}

\def\eqref#1{equation~\ref{#1}}

\def\1{\bm{1}}

\DeclareMathAlphabet{\mathsfit}{\encodingdefault}{\sfdefault}{m}{sl}
\SetMathAlphabet{\mathsfit}{bold}{\encodingdefault}{\sfdefault}{bx}{n}

\usepackage{color,xcolor}
\usepackage{epsfig}
\usepackage{graphicx}

\usepackage{adjustbox}
\usepackage{array}
\usepackage{booktabs}
\usepackage{colortbl}
\usepackage{wrapfig}
\usepackage{hhline}
\usepackage{multirow}
\usepackage{subcaption} %
\usepackage[font={small}]{caption}

\usepackage{amsmath,amsfonts,amssymb}
\usepackage{bm}
\usepackage{nicefrac}
\usepackage{microtype}

\usepackage{changepage}
\usepackage{extramarks}
\usepackage{fancyhdr}
\usepackage{lastpage}
\usepackage{setspace}
\usepackage{soul}
\usepackage{xspace}

\usepackage[pagebackref=true,breaklinks=true,colorlinks,citecolor=gray]{hyperref}

\usepackage{url}

\usepackage{algorithm, algorithmic}
\usepackage{enumerate}
\usepackage{todonotes} %
\usepackage{enumitem}  %
\usepackage{titlesec}
\usepackage{makecell}
\usepackage{pifont} %

\newcolumntype{L}[1]{>{\raggedright\let\newline\\\arraybackslash\hspace{0pt}}m{#1}}
\newcolumntype{C}[1]{>{\centering\let\newline\\\arraybackslash\hspace{0pt}}m{#1}}
\newcolumntype{R}[1]{>{\raggedleft\let\newline\\\arraybackslash\hspace{0pt}}m{#1}}

\newcommand{\sect}[1]{Sec.~\ref{#1}}

\newcommand{\sectapp}[1]{Appendix~\ref{#1}}
\newcommand{\eqn}[1]{Eq.~\ref{#1}}
\newcommand{\fig}[1]{Fig.~\ref{#1}}
\newcommand{\tbl}[1]{Table~\ref{#1}}

\newcommand{\ignore}[1]{}

\def\naive{na\"{\i}ve\xspace}

\makeatletter
\DeclareRobustCommand\onedot{\futurelet\@let@token\@onedot}
\def\@onedot{\ifx\@let@token.\else.\null\fi\xspace}

\def\eg{e.g\onedot} 
\def\ie{i.e\onedot}

\def\etal{et al\onedot}
\def\aka{a.k.a\onedot}
\makeatother

\definecolor{MyDarkBlue}{rgb}{0,0.08,1}
\definecolor{MyDarkGreen}{rgb}{0.02,0.6,0.02}
\definecolor{MyDarkRed}{rgb}{0.8,0.02,0.02}
\definecolor{MyDarkOrange}{rgb}{0.40,0.2,0.02}
\definecolor{MyPurple}{RGB}{111,0,255}
\definecolor{MyRed}{rgb}{1.0,0.0,0.0}
\definecolor{MyGold}{rgb}{0.75,0.6,0.12}
\definecolor{MyDarkgray}{rgb}{0.66, 0.66, 0.66}

\newcommand{\problemfull}{box program induction\xspace}
\newcommand{\Problemfull}{Box Program Induction\xspace}
\newcommand{\modelfull}{Box Program Induction\xspace}
\newcommand{\model}{BPI\xspace}
\newcommand{\program}{box program\xspace}
\newcommand{\Program}{Box Program\xspace}
\newcommand{\programs}{box programs\xspace}

\newcommand{\StructCompletion}{Huang~\etal\cite{Huang2014ImageCU}\xspace}
\newcommand{\SelfTuning}{Kaspar~\etal\cite{kaspar2015self}\xspace}

\newcommand{\myparagraph}[1]{\vspace{-4pt}\paragraph{#1}}
\newcommand{\myitem}{\vspace{-3pt}\item}

\titlespacing*{\section}{0pt}{0pt plus 2pt minus 2pt}{0pt plus 0pt minus 4pt}
\titlespacing*{\subsection}{0pt}{0pt plus 0pt minus 2pt}{0pt plus 0pt minus 2pt} 
\title{Multi-Plane Program Induction with 3D Box Priors}
\author{
Yikai Li$^{1,2*}$\quad Jiayuan Mao$^{1*}$\quad Xiuming Zhang$^{1}$\\
{\bf William T. Freeman$^{1,3}$\quad Joshua B. Tenenbaum$^{1}$\quad Noah Snavely$^{3}$\quad Jiajun Wu$^{4}$}\vspace*{5pt}\\
$^1$MIT CSAIL\qquad $^2$Shanghai Jiao Tong University\qquad\\
$^3$Google Research\qquad $^4$Stanford University\\
\vspace*{-20pt}
}

\begin{document}

\maketitle

\begin{abstract}
We consider two important aspects 
in understanding and editing images: modeling regular, program-like texture or patterns in 2D planes, and 3D posing of these planes in the scene. Unlike prior work on image-based program synthesis, which assumes the image contains a single visible 2D plane, we present \modelfull (\model), which infers a program-like scene representation that simultaneously models repeated structure on multiple 2D planes, the 3D position and orientation of the planes, and camera parameters, all from a single image. Our model assumes a {\it box prior}, i.e., that the image captures either an {\it inner view} or an {\it outer view} of a box in 3D. It uses neural networks to infer visual cues such as vanishing points or wireframe lines to guide a search-based algorithm to find the program that best explains the image. Such a holistic, structured scene representation enables 3D-aware interactive image editing operations such as inpainting missing pixels, changing camera parameters, and extrapolate the image contents.
\end{abstract}

\vspace{-5pt}
\section{Introduction}
\begin{figure}[t]
    \centering
    \includegraphics[width=\textwidth]{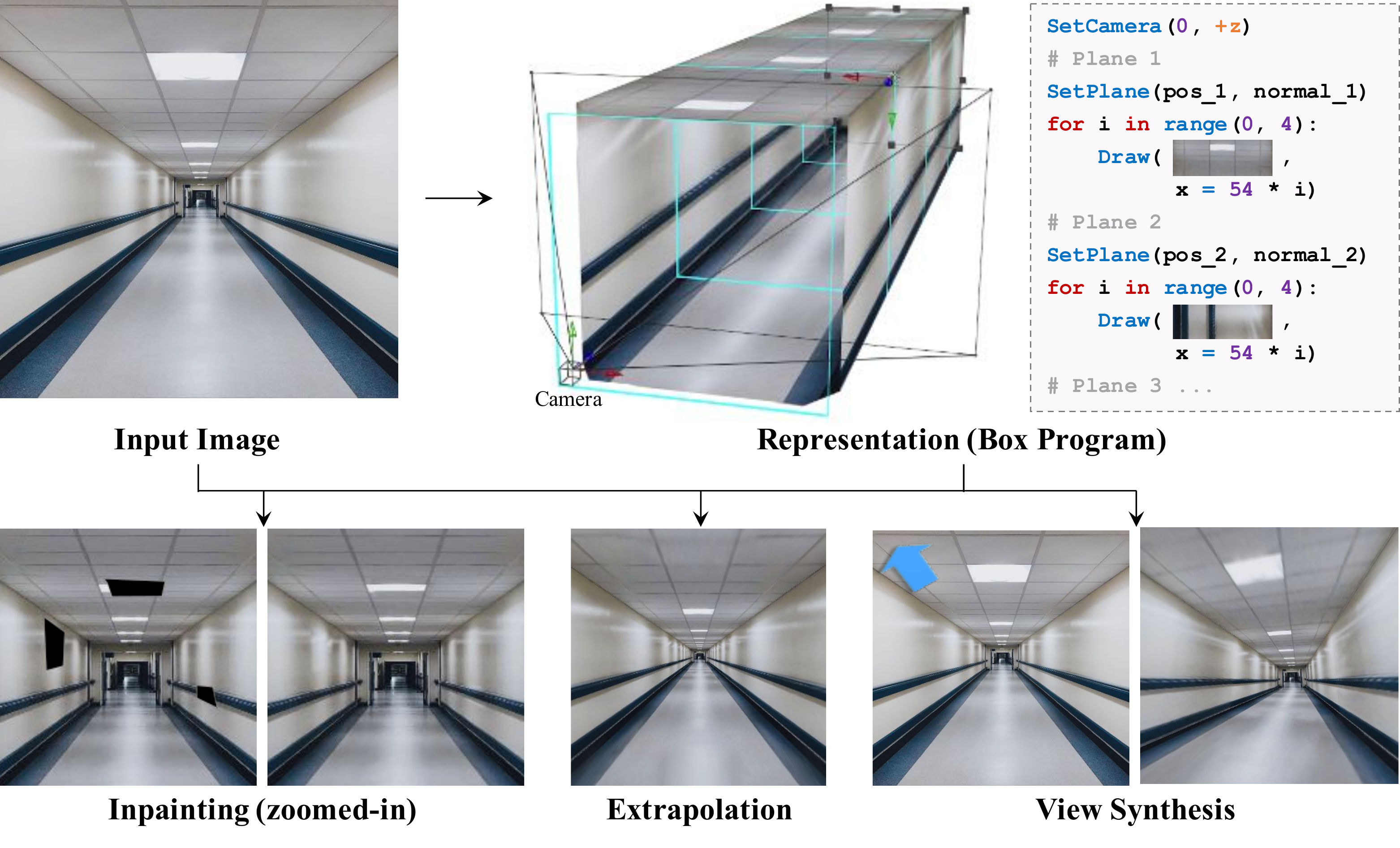}
    \vspace{-15pt}
    \caption{We present \modelfull (\model), which infers a program-like scene representation that simultaneously models repeated structure on {\it multiple} 2D planes, 3D positions and orientations of the planes, relative to the camera, all from a {\it single} image. The inferred program can be used to guide perspective- and regularity-aware image manipulation tasks,
    including image inpainting, extrapolation, and view synthesis. 
    }%
    \label{fig:teaser}
    \vspace{-10pt}
\end{figure}

We aim to build autonomous algorithms that can infer two important structures 
for compositional scene understanding and editing from a single image: the regular, program-like texture or patterns in 2D planes and the 3D posing of these planes in the scene.
As a motivating example, when observing a single image of a corridor like the one in \fig{fig:teaser}, we humans can effortlessly infer the camera pose, partition the image into five planes---including left and right walls, floor, ceiling, and a far plane---and recognize the repeated pattern on each of these planes. Such a holistic and structural representation allows us to flexibly edit the image, for instance by inpainting missing regions, moving the camera, and extrapolating the corridor to make it infinite.

A range of computer vision algorithms have utilized such a holistic scene representation to guide image manipulation tasks. Several recent ones fit into a program-guided image manipulation framework~\citep{ellis2018learning,liu2019learning,Mao2019Program}. These methods infer a program-like image representation that captures camera parameters and scene structures, enabling image editing operations guided by such programs so that the scene structure is preserved during editing. However, due to the combinatorial complexity of possible compositions of elementary components based on the program grammar, these methods usually only work for images in highly specific domains with a fixed set of primitives such as hand-drawn figures of simple 2D geometric shapes~\citep{ellis2018learning} and synthesized tabletop scenes~\citep{liu2019learning}, or natural images with of a {\it single} visible plane, such as ground tiles and patterned cloth~\citep{Mao2019Program,Li2020Perspective}.

To address these issues and scale up program-guided image manipulation, we present a new framework, namely, \modelfull (\model, for short), that jointly segments the image into multiple planes and infers the repeated structure on each plane. Our model assumes a {\it box prior}, leveraging the observation that box-like structures widely exist in images. Many indoor and outdoor scenes fall into this category: walking in a corridor or room corresponds to observing a box from the inside, and taking a picture of a building corresponds to seeing a box from the outside.

To enable efficient inference of box programs, we also propose to utilize mid-level cues, such as vanishing points or wireframe lines, as well as high-level visual cues such as subject segmentations, to implicitly constrain the search space of candidate programs. Given the input image, \model first infers these visual cues with pre-trained data-driven models.
Next, it enumerates all candidate programs that satisfy the implicit constraints imposed by these inferred visual features. Finally, it ranks all candidate programs using low-level visual features such as pixel reconstruction.

In summary, we present \model, a framework for inducing {\it \programs} from images by exploiting learned visual cues. Our experiments show that \model can efficiently and accurately infer the structure and camera parameters for both indoor and outdoor scenes. 
The inference procedure is robust to errors and noise inherent to visual cue prediction: \model automatically selects the best candidate wireframe lines and refines the vanishing points if they are not accurate.
\model also enables users to make 3D-aware interactive editing to images, such as inpainting missing pixels, extrapolating image content in specific directions, and changing camera parameters.

\section{Related Work}

\paragraph{Visual program induction.}
Computer graphics researchers have used procedural modeling (mostly top-down) for representing 3D shapes~\citep{li2017grass,tian2019learning} and indoor scenes~\citep{wang2011symmetry,li2019grains,niu2018im2struct}.
With advances in deep networks, some methods have paired top-down procedural modeling with bottom-up recognition networks. Such hybrid models have been applied to hand-drawn sketches~\citep{ellis2018learning}, scenes with simple 2D or 3D geometric primitives~\citep{sharma2018csgnet,liu2019learning}, and markup code~\citep{deng2017image,beltramelli2018pix2code}. The high-dimensional nature of procedural programs poses significant challenges to the search process; hence, even guided by neural networks, these works focus only on synthetic images in constrained domains.
SPIRAL~\citep{ganin2018synthesizing} and its follow-up SPIRAL++~\citep{mellor2019unsupervised} use reinforcement learning to discover latent ``doodles'' that are later used to compose the image. Their models work on in-the-wild images, but cannot be directly employed in tasks  involving explicit reasoning, such as image manipulation and analogy making, due to the lack of program-like, interpretable representations. %

In the past year, \citet{young2019learning} and \citet{Mao2019Program} integrated formal programs into deep generative networks to represent natural images, and later applied the hybrid representation to image editing. \citet{Li2020Perspective} extended these models by jointly inferring perspective effects. All these models, however, assume a single plane in an image, despite the fact that most images contain multiple planes such as floor and ceiling. Our \model moves beyond the single-plane assumption by leveraging box priors.

\myparagraph{Image manipulation.}

Image manipulation, in particular image inpainting, is a long-standing problem in graphics and vision.
Pixel-based~\citep{ashikhmin2001synthesizing,ballester2001filling} and patch-based methods~\citep{efros2001image,barnes2009patchmatch,he2012statistics} achieve impressive results for inpainting regions that requires only local, textural information. They fail on cases requiring high-level, structural, or semantic information beyond textures. Image Melding~\citep{Darabi2012ImageMC} extends patch-based methods by allowing additional geometric and photometric transformations on patches, but ignores global consistency among patches. Huang~\etal\cite{Huang2014ImageCU} also use perspective correction to assist patch-based inpainting, but rely on vanishing points detected by other methods. In contrast, \model segments planes and estimates their normals based on the global regularity of images.

Advances in deep networks have led to impressive inpainting algorithms that integrate information beyond local pixels or patches~\citep{iizuka2017globally,yang2017high,yu2018generative,liu2018image,yu2019free,zhou2018non,yan2018shift,xiong2019foreground,nazeri2019edgeconnect}.  In particular, deep models that learn from a single image (\aka, internal learning) produce high-quality results for image manipulation tasks such as inpainting, resizing, and expansion~\citep{shocher2019ingan,zhou2018non,shaham2019singan}. \model also operates on a single image, simultaneously preserving the 3D structure and regularity during image manipulation.
\section{\Program Induction}
Our proposed framework, \modelfull (\model), takes an image as input and infers a \program that best describes the image, guided by visual cues. 

\begin{figure}[t]
    \centering
    \includegraphics[width=\linewidth]{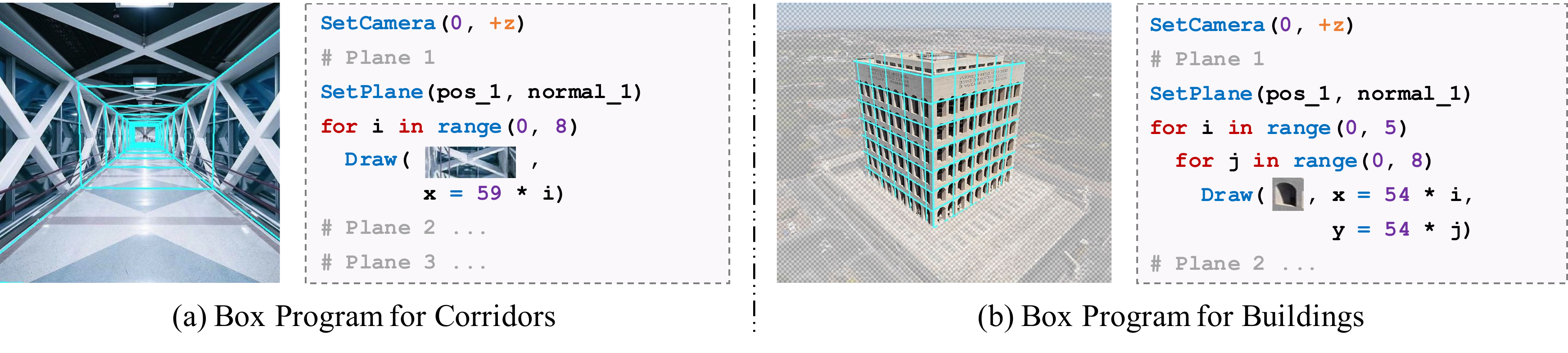}
    \vspace{-20pt}
    \caption{Example \programs inferred by \model. Our \program jointly models camera parameters, the 3D positions and orientations of multiple planes, as well as the regularity structure on individual planes.}
    \vspace{-5pt}
    \label{fig:program_examples}
\end{figure}  
\subsection{Domain-Specific Language}
\begin{table}[t]
\centering\small
  \begin{minipage}[c]{0.77\textwidth}
    \setlength{\tabcolsep}{0pt}
    \centering
    \begin{tabular}{rcp{0.71\columnwidth}}
    \toprule
        Program & $\longrightarrow$ & CameraProgram; WorldProgram;\\
        CameraProgram & $\longrightarrow$ & SetCamera(pos={\tt Vec3}, point\_to={\tt Vec3});\\
        WorldProgram  & $\longrightarrow$ & PlaneProgram; $\mid$ PlaneProgram; World Program; \\
        PlaneProgram & $\longrightarrow$ & SetPlane(pos={\tt Vec3}, normal={\tt Vec3})\; For1Stmt; \\
        For1Stmt & $\longrightarrow$ & {\tt For} (~$i$~{\tt in} {\tt range}(Integer, Integer) )\{ DrawStmt; $\mid$ For2Stmt \}\\
        For2Stmt & $\longrightarrow$ & {\tt For} (~$j$~{\tt in} {\tt range}(Integer, Integer) )\{ DrawStmt \}\\
        DrawStmt & $\longrightarrow$ & {\tt Draw} ($x$=Expr, $y$=Expr) \\
        Expr & $\longrightarrow$ & $\text{\tt Real} \times i + \text{\tt Real} \times j \mid \text{\tt Real} \times i$\\
    \bottomrule
    \end{tabular}
  \end{minipage}\hfill
  \begin{minipage}[c]{0.23\textwidth}
    \caption{The domain-specific language~(DSL) of \programs. Language tokens {\tt For}, {\tt If}, {\tt Integer}, {\tt Real}, and arithmetic/logical operators follow the Python convention. {\tt Vec3} denotes 3D real vectors.}
    \label{tab:dsl}
  \end{minipage}
  \vspace{-10pt}
\end{table} 
\label{sec:dsl}
We start by describing the domain-specific language (DSL) that we use to express the multiple planes in the scene and the regularity structure on individual planes, namely the \textit{\programs}. We assume these planes are faces of a 3D box. Take \fig{fig:program_examples}a as an example: the corridor is composed of four planes, each containing visually regular patterns.

\tbl{tab:dsl} specifies the DSL, and \fig{fig:program_examples} shows sample programs. A \program consists of two parts: camera parameters and programs for individual planes. A {\it plane program} first sets the plane's surface normal, then specifies a sub-program defining the regular 2D pattern on the plane. These patterns utilize the primitive {\tt Draw} command, which places patterns at specified 2D positions. {\tt Draw} commands appear in {\tt For}-loop statements that characterize the regularity of each plane.

\subsection{\Program Fitness}
\label{sec:low-level}
Given an input image, our goal is to segment the image into different planes, estimate their surface normals relative to the camera, and infer the regular patterns. We treat this problem as seeking a program $P$ that best fits the input image $I$. We first define the {\it fitness} of a program $P$ by measuring how well $P$ reconstructs $I$.
Recall that a \program $P$ is composed of multiple {\it plane programs}, each of which defines the regular pattern on a plane as well as its position and orientation 
in 3D. Combined with camera parameters, we can use parameters of each plane program to {\it rectify} the corresponding plane, resulting in images without perspective effects $\{J_1, J_2, \cdots, J_k\}$, where $k$ is the number of planes described 
by $P$.

For each plane $i$, we compute its fitness score by comparing its rectified image $J_i$ and the corresponding plane program block, denoted as $Q_i$. The fitness score is defined in a similar way as in the Perspective Plane Program Induction framework~\citep{Li2020Perspective}. Specifically, executing $Q_i$ produces a set of 2D coordinates $C_i$ that can be interpreted as the centers of each visual element, such as the center of the lights on the ceiling in \fig{fig:program_examples}a. Since $Q_i$ contains a nested loop of up to two levels, we denote the loop variable for each {\tt For} loop as $a$ and $b$. Thus, each 2D coordinate in $C_i$ can be written as a function of $a$ and $b$, $c(a,b)\in\mathbb{R}^2$. The fitness score $F$ is defined as

{
\vspace{-10pt}
\small
\begin{adjustwidth}{0em}{0em}
\begin{equation}
F = -\sum_{a, b} \left[\left\|J_i[c(a, b)] - J_i[c(a+1, b)] \right\|_2^2 + \left\|J_i[c(a, b)] - J_i[c(a, b+1)] \right\|_2^2\right]\label{eq:loss},
\end{equation}
\end{adjustwidth}
\vspace{-10pt}
}%
where $\|\cdot\|_2$ is the $\mathcal{L}_2$ norm and $J_i[c(a,b)]$ denotes the patch centered at 2D coordinates $c(a,b)$. Since we only consider lattice patterns on individual planes, we implement this by first shifting $J_i$ leftward (or downward) by the `width' (or `height') of a lattice cell and then computing the pixel-wise difference between the shifted image and the original image $J_i$. The overall program fitness of $P$ is the sum of the fitness function for all planes.

\begin{figure}[t]
    \centering
    \includegraphics[width=\linewidth]{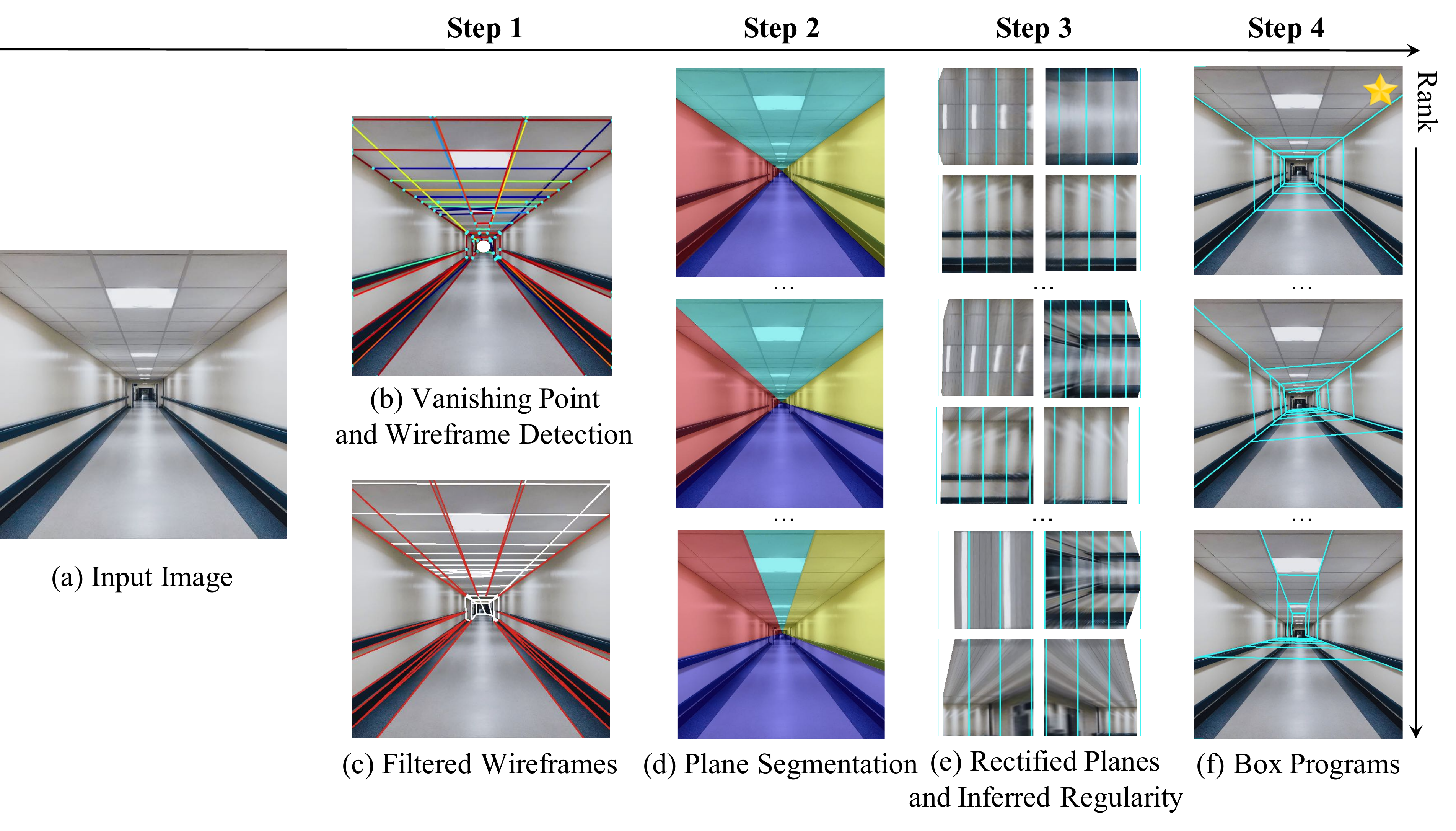}
    \vspace{-15pt}
    \caption{Our \modelfull finds the best-fit program that describes the input image (a). It first detects the vanishing point and wireframe segments (b), followed by a filtering step (c). It then constructs a set of candidate plane segmentation maps (d). Given each plane segmentation, it rectifies each plane and infers its regularity structure (e). We rank all candidate \programs by their fitness (f); the starred candidate is the best.
    }
    \vspace{-10pt}
    \label{fig:model}
\end{figure} 
\subsection{3D Box Priors and the Role of Visual Cues}
\label{sec:mid-high-level}

A \naive way to find the best-fit program $P$ is to enumerate all possible plane segmentations of the input image $I$ %
and rank all candidates by the fitness score (\eqn{eq:loss}). However, the complexity of this \naive method scales exponentially in the number of planes.
Instead, we propose to consider first, the {\it box prior}, which constrains the plane segmentation of the image and, second, visual cues that help to guide the search. Specifically, we impose the following {\it box prior}:
\begin{itemize}[leftmargin=*]
    \myitem For an {\it inner view} of a box, \eg, a corridor as in \fig{fig:program_examples}a, our \program models four planes: two side walls, the floor, and the ceiling. For images with a far plane, we will segment the far plane but do not use programs to model it, as most far planes do not contain a regular structure. 
    \myitem For an {\it outer view} of a box, \eg, a building as in \fig{fig:program_examples}b, our \program models the two side walls as two planes. We do not model the roof as, in most images, the roof is either nonvisible or very small in size. %
\end{itemize}
\vspace{-5pt}

Below, we show how to use {\it visual cues} to guide the search for \programs. We focus on the {\it inner view} case, and include details for the {\it outer view} case in the supplemental material. The full search process is illustrated in \fig{fig:model}, and consists of four steps. First, we use pre-trained neural networks to detect the vanishing point and the 2D wireframes line segments from the image. We also filter out invalid wireframe segments. Next, we generate a set of candidate plane segmentations based on the detected wireframe segments. Then, for each segmented plane, we seek the program that best describes the plane structure. Finally, we rank all candidate plane segmentations by the fitness score.

\myparagraph{Step 1: Visual cue detection.}
Following the {\it box prior}, all inner views of a box contain a single {\it vanishing point} and four {\it intersection lines} that are the intersection between the two walls, the ceiling, and the floor in 3D. These intersection lines will be projected onto the image plane as four lines that intersects at the vanishing point. We use this property to constrain the candidate plane segmentation. Leveraging vanishing point information for inferring box structures of scenes has also been studied in~\cite{hedau2009recovering}.

Given an input image (\fig{fig:model}a), we apply NeurVPS~\citep{zhou2019neurvps} to detect the vanishing point and L-CNN~\citep{zhou2019end} to extract wireframes in the image. We use the most confident prediction of NeurVPS as the vanishing point $\mathit{vp} \in \mathbb{R}^2$, which is a 2D coordinate on the image plane. Each wireframe segment is represented as a segment $AB$ on the image, from $(x^A, y^A)$ to $(x^B, y^B)$, as illustrated in \fig{fig:model}b. Next, we filter out wireframe segments whose length is smaller than a threshold $\delta_{1}$ or whose extension does not cross a neighbourhood centered at $\mathit{vp}$ with radius $\delta_2$. The remaining wireframe segments, denoted by the set $\mathit{WF}$, are illustrated in \fig{fig:model}c.

\myparagraph{Step 2: Plane segmentation.}
We then enumerate all combinations of four wireframe segments from $\mathit{WF}$. As these wireframe segments $s_i = [(x_i^A, y_i^A), (x_i^B, y_i^B)], i=1,2,3,4$ may not intersect at a single point, we compute a new vanishing point $\mathit{vp}^*$ that minimizes $\sum_{i} \mathit{dist}(\mathit{vp}^*, s_i)$, where $\mathit{dist}$ is the distance between the point $\mathit{vp}^*$ and the line containing segment $s_i$. Next, we connect the new vanishing point $\mathit{vp}^*$ and the farther end of each $s_i$ to get four rays. These four rays partition the image into four parts, which we treat as the segmentation of four planes, as illustrated in \fig{fig:model}d.

\myparagraph{Step 3: Plane rectification and regularity inference.}
Fixing the camera at the world origin, pointing in the +$z$ direction, we then compute the 3D position and surface normal of each plane. As shown in \fig{fig:teaser}, because the distance between camera and the corridor is coupled with the focal length of the camera, here we use a fixed focal length of $f=35 \mathrm{mm}$\footnote{Following common practice, we also fix other camera intrinsic properties: optical center to $(0, 0)$, skew factor to $0$, and pixel aspect ratio to $1$.}. Based on these assumptions, the four rays can {\it unambiguously} determine the 3D positions and surface normals of four planes. The proof can be found in the supplemental material.

Based on the inferred surface normal, we can {\it rectify} each plane, yielding a set of rectified images $\{J_1, J_2, \cdots, J_4\}$. For each rectified plane, we search for the best plane program that describes it, based on the fitness function \eqn{eq:loss}. The inferred plane regularity structures are shown in \fig{fig:model}e.

\myparagraph{Step 4: Box program ranking.}
We sum up the fitness score for four planes in each candidate segmentation as the overall program fitness. We use this score to rank all candidate segmentations, and choose the program with the highest fitness to describe the entire image.

\begin{figure}[t]
    \centering
    \includegraphics[width=\linewidth]{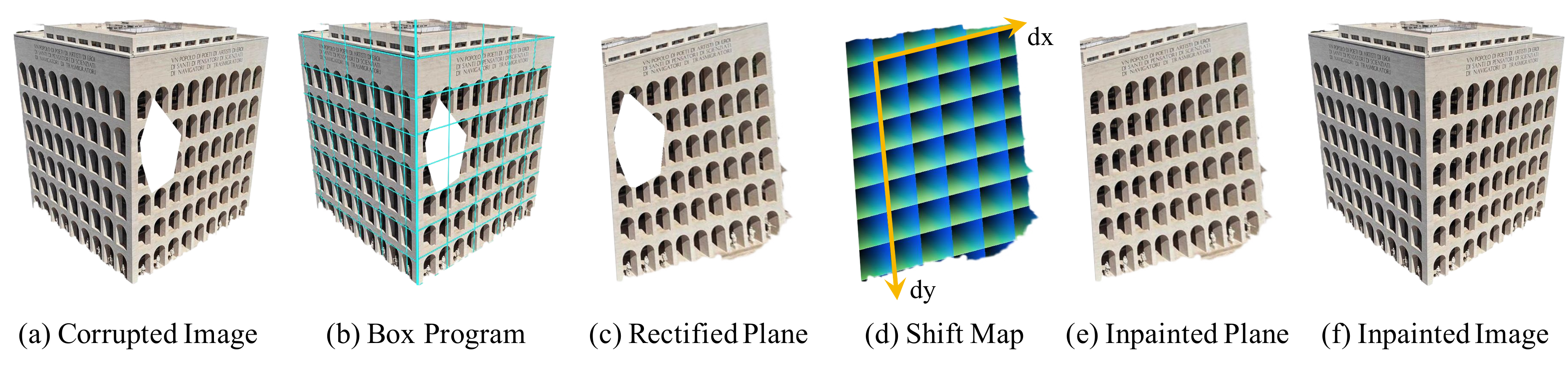}
    \vspace{-15pt}
    \caption{An illustration of the proposed program-guided PatchMatch. Given the corrupted image, we first detect its \program (b) and rectify each plane (c). We use the regular structure (d) on the plane to guide the PatchMatch to inpaint the corrupted plane (e). The color in (d) visualizes  the relative position of each pixel to the center of its associated lattice cell.}
    \vspace{-15pt}
    \label{fig:algorithm}
\end{figure} 
\subsection{Program-Guided Image Manipulation}
\label{sec:use}
The inferred \program enables 3D-aware interactive image manipulation: inpainting missing pixels, changing camera parameters 
, and extrapolating the image content. Lying at the core of these applications is a program-guided PatchMatch algorithm.

Our program-guided PatchMatch augments the PatchMatch algorithm~\citep{barnes2009patchmatch} by enforcing the regularity on each plane. Consider the task of inpainting missing pixels on the wall of a building in \fig{fig:algorithm}. Taking a corrupted image as input (\fig{fig:algorithm}a), we infer the \program from the undamaged pixels (\fig{fig:algorithm}b). The inferred \program explains the plane segmentation and the regularity pattern on each plane. For each plane that contains missing pixels, we can first rectify the corrupted plane as shown in \fig{fig:algorithm}c. Since the plane program on this plane will partition the image with a lattice, we construct a ``shift map'', illustrated as \fig{fig:algorithm}d, which represents the relative position of each pixel to the center of the lattice cell that this pixel lies in, normalized to $[-0.5, 0.5]$.

In PatchMatch, the similarity $\mathit{sim}(p, q)$ between pixel $p$ and pixel $q$ is computed as a pixel-level similarity $\mathit{sim}_{\mathrm{pixel}}$ between two patches centered at $p$ and $q$, with a constant patch size $\delta_{\mathrm{pm}}$. We add another term to this similarity function:
\begin{equation}
    \mathit{sim}(p, q) \triangleq \mathit{sim}_{\mathrm{pixel}} + \mathit{sim}_{\mathrm{reg}} = \mathit{sim}_{\mathrm{pixel}} - \lambda_{\mathrm{reg}} \| \mathrm{wraparound}(\mathit{smap}[p] - \mathit{smap}[q]) \|^2_2,
\end{equation}
where $\lambda_{\mathrm{reg}}$ is a hyperparameter that controls the weight of the regularity enforcement. $\mathrm{abs}$ is the absolute value function.
The shift-map term measures whether two pixels $p$ and $q$ correspond to the same location on two (possibly different) repeating elements on the plane. 
We ``wrap around'' shift map distance by $\mathrm{wraparound}(\mathbf{x}) \equiv \max(1-\mathbf{x}, \mathbf{x})$, as the top-left corner and the bottom-right corner of a cell also matches. 
Thus, the PatchMatch algorithm will choose the pixel based on both pixel similarity and regularity similarity to fill in the missing pixels (\fig{fig:algorithm}e).
\section{Experiments}

For evaluation, we introduce two datasets,
and then apply \programs to vision and graphics tasks on these datasets, including plane segmentation (\sect{sect:result_plane_segmentation}), image inpainting and extrapolation (\sect{sect:result_inpaint}), and view synthesis (\sect{sect:result_viewsyn}). 

\myparagraph{Dataset.}

We collect two datasets from web image search engines for our experiments, a 44-image {\it Corridor Boxes} dataset and a 42-image {\it Building Boxes} dataset. These correspond to the inner view and the outer view of boxes, respectively.
For both datasets, we manually annotate the plane segmentations by specifying edges of the boxes. For corridor images, we also create a mask for the far plane.
For building images, we supplement the subject segmentation (\ie, the building of interest) to the dataset annotation. 

\subsection{Plane Segmentation}
\label{sect:result_plane_segmentation}

Because \model develops a 3D understanding of the scene in the process of inferring the
\program under perspective effects, an immediate application 
is single-image plane segmentation. Note that \model performs this task by jointly inferring the perspective effects and the pattern regularity, in contrast to supervised learning approaches that train models with direct supervision, such as~\cite{Liu_2019_CVPR}. The core intuition is that the correct plane segmentation leads to the program 
that best describes each segmented plane in terms of repeated visual elements.

\myparagraph{Baselines.}
We compare our method with two baselines: \StructCompletion and PlaneRCNN~\citep{Liu_2019_CVPR}.
\StructCompletion first uses VLFeat~\citep{vedaldi08vlfeat} to detect vanishing points and line segments. It then generates a soft partition of the image into multiple parts by the support lines. PlaneRCNN is a learning-based method for plane segmentation. It uses a Mask-RCNN-like pipeline~\citep{he2017mask} and is trained on ScanNet~\citep{dai2017scannet}.

\begin{wraptable}{r}{.45\linewidth}
    \small \centering
    \vspace{-1em}
    \setlength\tabcolsep{3pt}
    \begin{tabular}{lcccc}
        \toprule
        Method & CrdO & CrdC & BldO & BldC \\
        \midrule
        \StructCompletion & 0.30 & 0.30 & 0.73 & 0.69 \\
        PlaneRCNN & 0.52 & 0.52 & 0.63 & 0.63 \\
        \model (Ours) & {\bf 0.84} & {\bf 0.84} & {\bf 0.86} & {\bf 0.86} \\
        \arrayrulecolor{black}\bottomrule
    \end{tabular}
    \vspace{-5pt}
    \caption{Plane segmentation results in IoU between detected and groundtruth planes. \model outperforms both baselines on both original (CrdO, BldO) and corrupted images (CrdC, BldC).}
    \label{tab:result_partition}
    \vspace{1em}
\end{wraptable} 
\myparagraph{Metrics.}
The output of each model is a set of masks indicating the segmented planes. We compare these with the ground-truth masks by computing the best match between two sets of masks, where the matching metric is the intersection over union (IoU).
We then report the average IoU of all predicted masks. For corridor images, we exclude pixels in the far plane region during evaluation. Because we wish to use the plane segmentation to aid in image manipulation tasks such as image inpainting, we evaluate all methods on both original images (CrdO, BldO) and corrupted images (CrdC, BldC).

\begin{figure}[t]
    \centering
    \includegraphics[width=\linewidth]{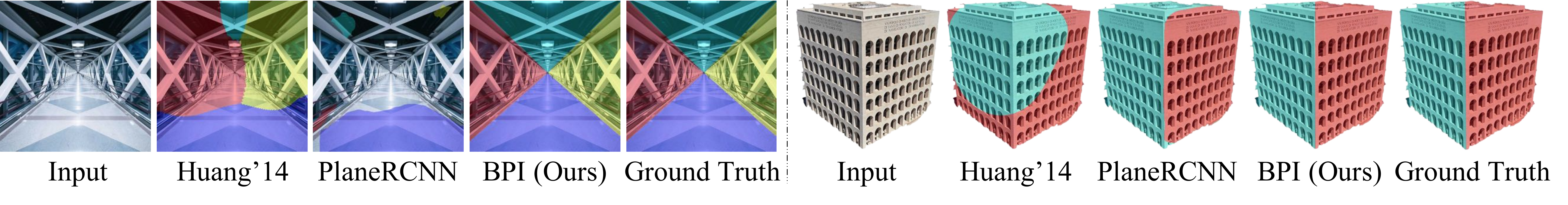}
    \vspace{-15pt}
    \caption{Visualization of the plane segmentation by different methods.}%
    \vspace{-5pt}
    \label{fig:qresult_building_partition}
\end{figure} 
\myparagraph{Results.}
\fig{fig:qresult_building_partition} and \tbl{tab:result_partition} show that \model consistently outperforms the baselines on both corridor and building images. The baselines fail to detect planes when they contain complex structures and patterns. %

\subsection{Image Inpainting}
\label{sect:result_inpaint}

The inferred \programs support 3D-aware and regularity-preserving image manipulation, because they provide information on both what the perspective effects are and how the visual element repeats. To test \model's performance on such tasks, we generate a dataset of corrupted images by randomly masking out regions of images from 
our two datasets.

\myparagraph{Metrics.}
We use four metrics: pixel-level $\mathcal L_1$ distance, peak signal-to-noise ratio (PSNR),  structural similarity index (SSIM)~\citep{wang2004image}, and a learned perceptual metric LPIPS~\citep{zhang2018unreasonable}. Following standard practice~\citep{merkle2007multi, balle2019end}, we compute PSNR and SSIM on image luma, and compute $\mathcal L_1$ distance and LPIPS directly on RGB values.

\myparagraph{Baselines.}
We compare our model against both learning-based GatedConv~\citep{yu2019free} and three non-learning-based algorithms: PatchMatch~\citep{barnes2009patchmatch}, Image Melding~\citep{Darabi12:ImageMelding12}, and \StructCompletion. All three non-learning algorithms are patch-based. Image Melding allows additional geometric and photometric transformations on patches. \StructCompletion first segments the image into different planes and augments a standard PatchMatch procedure with the plane rectification results.

\begin{table}[t]
    \small \centering
    \setlength\tabcolsep{3.5pt}
    \begin{tabular}{lcccccccc}
        \toprule
        & \multicolumn{4}{c}{Corridors} & \multicolumn{4}{c}{Buildings} \\
        \cmidrule(lr){2-5}\cmidrule(lr){6-9}
        Method & $\mathcal{L}_1$ Mean $\downarrow$ & PSNR $\uparrow$ & SSIM $\uparrow$ & LPIPS $\downarrow$ & $\mathcal{L}_1$ Mean $\downarrow$ & PSNR $\uparrow$ & SSIM $\uparrow$ & LPIPS $\downarrow$ \\
        \midrule
        PatchMatch & 53.12 & 31.55 & 0.9872 & 0.0215 & 34.88 & 32.99 & 0.9896  & 0.0072 \\
        Image Melding & 58.50  & 30.54 & 0.9880 & 0.0174 & 49.10 & 30.42 & 0.9890 & 0.0134 \\
        \StructCompletion & 51.88 & 31.41 & 0.9869 & 0.0177 & {\bf 26.10} & {\bf 34.87} & {\bf 0.9917} & {\bf 0.0049} \\
        GatedConv & 44.98 & 32.32 & {\bf 0.9883} & {\bf 0.0153} & 55.31 & 29.54 & 0.9866  & 0.0112 \\
        \model (Ours) & {\bf 38.45} & {\bf 34.21} & {\bf 0.9892} & {\bf 0.0170} & {\bf 26.43} & {\bf 34.86} & {\bf 0.9913} & {\bf 0.0054} \\
        \bottomrule
    \end{tabular}
    \vspace{5pt}
    \caption{We compare \model-guided PatchMatch with both patch-based and learning-based methods on the task of image inpainting. $\uparrow$ indicates that the higher the number, the better. {\bf Bold} indicates models that are indistinguishable with the best one under that metric with a linear mixed model. See text for details. }
    \label{tab:result_inpaint}
    \vspace{-10pt}
\end{table} %
\begin{figure}[t]
    \centering
    \includegraphics[width=\linewidth]{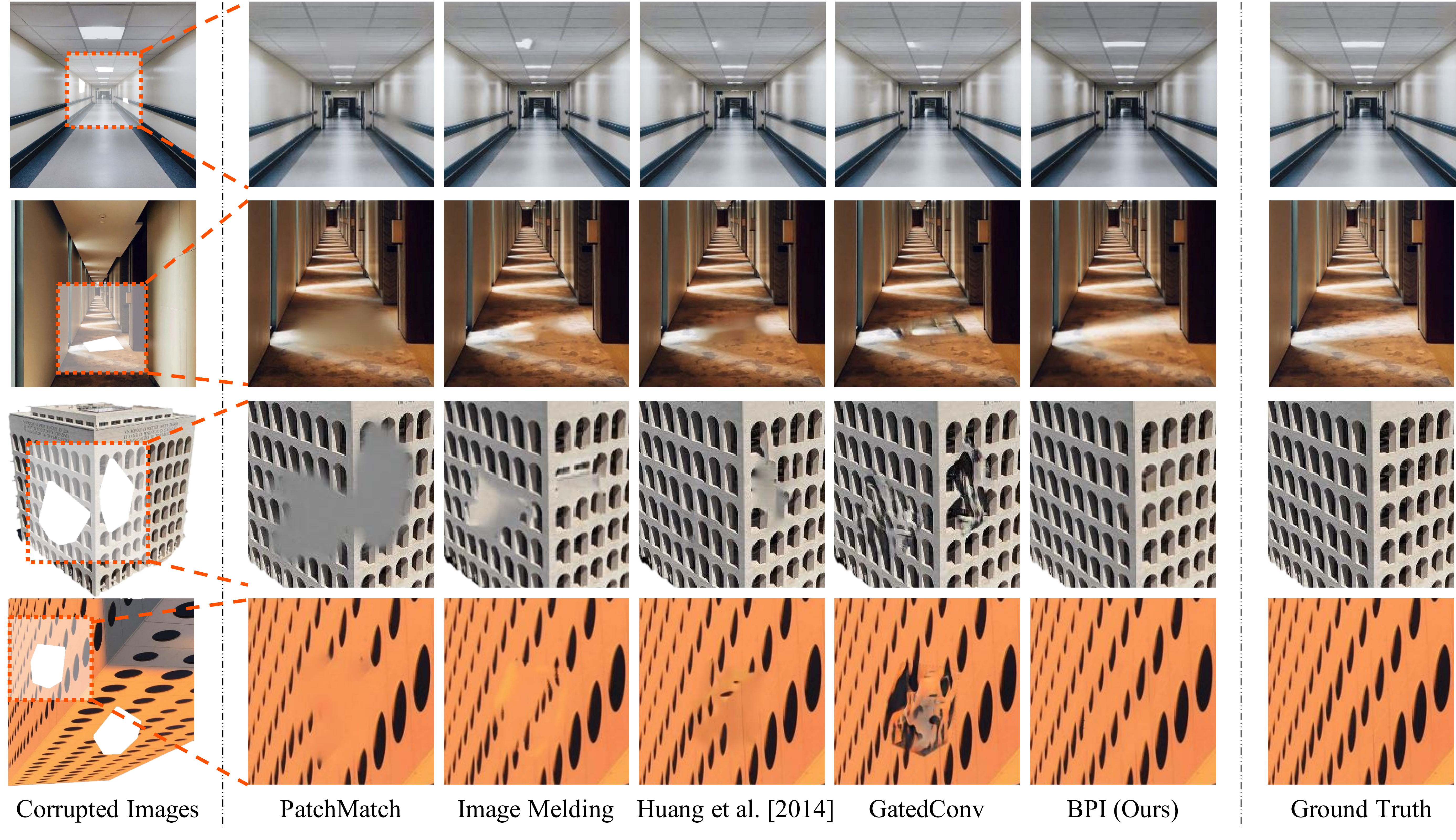}
    \vspace{-15pt}
    \caption{Qualitative results on the task of image inpainting. Compared with the baselines, our model can better preserve the regular structures even if they are sparse or have strong perspective effects.
    }
    \vspace{-10pt}
    \label{fig:qresult_building_inpaint}
\end{figure} 
\myparagraph{Results.}
The results are summarized in \tbl{tab:result_inpaint} and \fig{fig:qresult_building_inpaint}. Here we run linear mixed models between each pair of methods and, for each metric, we mark in bold all methods that are indistinguishable with the best one. All $p$-values are in the supplementary material. Our method outperforms baselines on corridor images and achieves comparable results on building images with \StructCompletion. As discussed in \sect{sect:result_plane_segmentation}, \StructCompletion relies on straight lines on the plane to segment and rectify the image, so it works well on planes with a plethora of such features. %
\StructCompletion tends to fail on images without dense straight lines on the plane (rows 3-4 of \fig{fig:qresult_building_inpaint}). On corridor images, beyond producing high-fidelity inpainting results, our regularity-aware PatchMatch process preserves the structure of the scene, such as the light on the ceiling in row 1 of \fig{fig:qresult_building_inpaint}. 

\myparagraph{Image extrapolation.}
\begin{figure}[t]
    \centering
    \includegraphics[width=\linewidth]{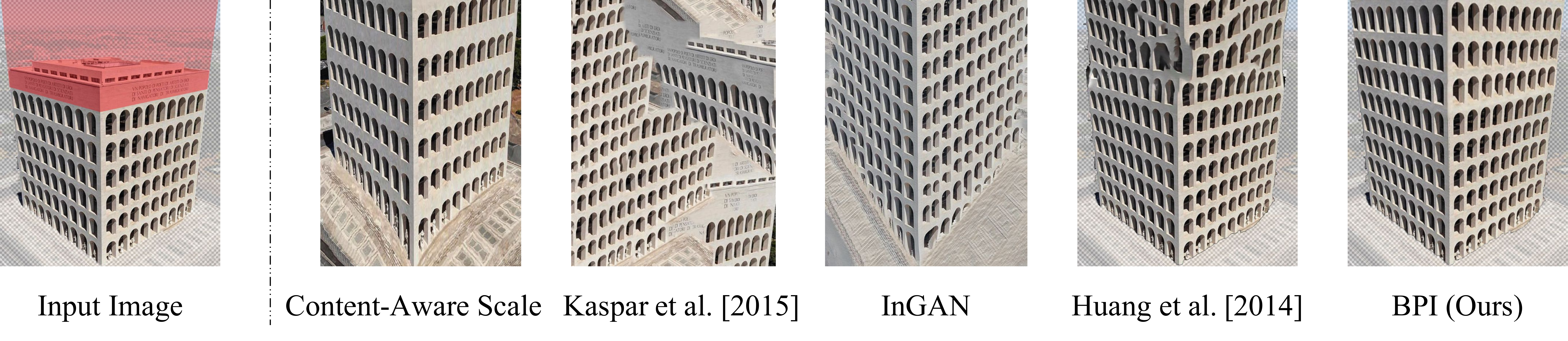}
    \vspace{-20pt}
    \caption{The ``extrapolated'' buildings. Content-Aware Scale, \SelfTuning, and InGAN fail to preserve the building structure while extrapolating the image: they either generate irregular patterns or change the shape of the planes. \StructCompletion fails to preserve the regular structure when inpainting large areas.}
    \vspace{-10pt}
    \label{fig:qresult_extrapolation}
\end{figure} 

Beyond inpainting missing pixels, our regularity-aware algorithm can extrapolate the box structure. Here, on the {\it Building Boxes} dataset, we show that our model can make the building taller or wider. The input to the model is a foreground mask of the building and the target region to be filled with the extrapolated building. We compare our method with four baselines: Content-Aware Scaling in Adobe Photoshop, \SelfTuning, \StructCompletion and InGAN~\citep{shocher2019ingan}. For content-aware scaling, we first select the foreground mask and then scale it so that it fills the target region. For both \SelfTuning and InGAN, we extract the bounding box of the foreground building and use it as the input. The model generates a new image that is 1.5x larger. For \StructCompletion, we cast the extrapolation problem as inpainting the target region.

As shown in \fig{fig:qresult_extrapolation}, Content-Aware Scaling is unaware of perspective effects and fails to preserve the lattice structure in the image. It also cannot generate new visual elements such as windows. Both \SelfTuning and InGAN do not preserve existing pixels when extrapolating the image. \SelfTuning, as a texture synthesis method, also ignores the two-plane structure when generating the new image. While InGAN is able to capture the visual elements, it does not follow the lattice pattern during synthesis. \StructCompletion can inpaint small local areas as in \fig{fig:qresult_building_inpaint}, but does not follow the regular structure when inpainting a large area. In contrast, \model preserves both the plane structure of the building and the lattice pattern on each side.

Quantitatively, we randomly select 12 images, and ask 15 people to rank the outputs of different methods. We collected $12 \times 15 = 180$ responses. The preference for the models are: Ours($61\%$), Content-Aware Scaling($16\%$), \SelfTuning($2\%$), InGAN($16\%$), and \StructCompletion($5\%$).

\subsection{View Synthesis}
\label{sect:result_viewsyn}

\begin{figure}[t]
    \centering
    \includegraphics[width=\textwidth]{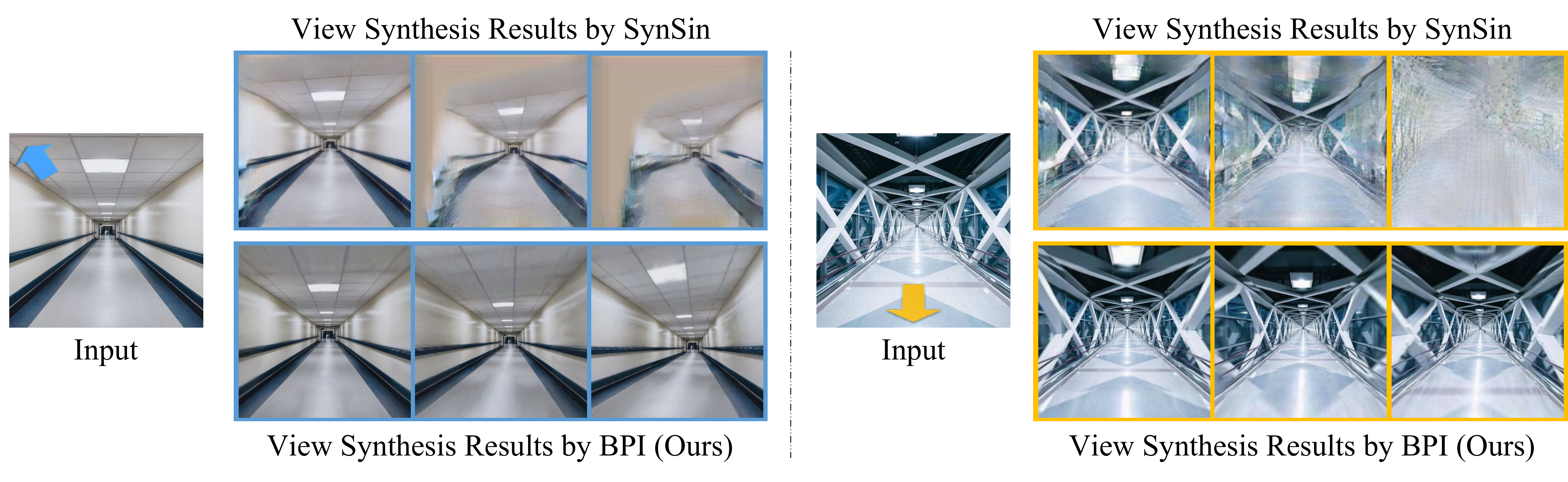}
    \vspace{-20pt}
    \caption{View synthesis from a single image of a corridor. Compared with the learning-based method SynSin, our model better preserves the regular patterns on the walls and also has remarkably fewer artifacts.}%
    \label{fig:qresult_view_synthesis}
\end{figure} 
As our inferred \programs captures a holistic 3D scene representation, 
we can synthesize novel views of the scene from a {\it single} image.
We compare different models on the {\it Corridor Boxes} dataset. 
We consider three types of camera movement: 1) ``step into the corridor'', 2) ``step back in the corridor'', and 3) ``step back in the corridor, pan leftward, and tilt upward''. All trajectories generate 5 frames. Detailed parameters are included in the supplemental material.
Note that in both trajectories that involve ``stepping back'', the algorithm must synthesize pixels unseen in the original image. For our \model, we run our program-guided PatchMatch to extrapolate the planes and synthesize pixels that are outside the input view frustum.

\myparagraph{Results.} 
We compare images generated by our \model and by SynSin~\citep{wiles2019synsin} in \fig{fig:qresult_view_synthesis}. The arrow on the input shows camera movement. As our method can generate a corridor of an arbitrary length, we see significantly fewer artifacts when the camera movement is large, compared with SynSin. Note that even in the first columns where the camera movement is small, the pixels synthesized by SynSin that are outside the original view frustum already fail to preserve the regular structure.

For a quantitative comparison, we randomly select 20 images, generate synchronized videos of the results produced by our method and by SynSin on all three camera trajectories, and ask 10 people to rank the outputs.
We collected $20 \times 3 \times 10 = 600$ responses. For the three different trajectories, $100\%$, $94\%$, and $99.5\%$ of the responses prefer our result to that by SynSin, respectively.

\subsection{Failure Case}

\begin{figure}[t]
    \centering
    \includegraphics[width=0.8\textwidth]{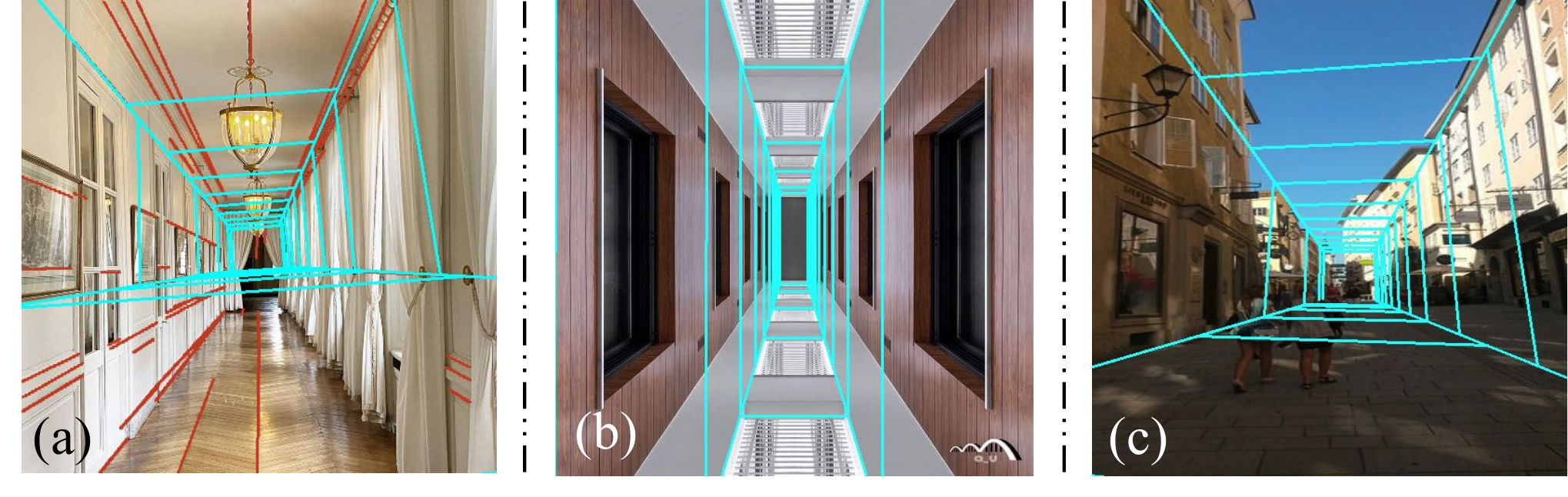}
    \caption{Failure cases. Our model may fail when neural networks misdetect visual cues (a).
    When image contains a solid color plane, our model may segment the pure white part to other planes (b). The inference might fail on irregular scenes, but could be mitigated with user interaction (c).}
    \vspace{-10pt}
    \label{fig:qresult_failure_case}
\end{figure} 
\fig{fig:qresult_failure_case} shows three main failure cases of our model. First, our model might misdetect vanishing points and wireframe segments, as illustrated in (a), where the model missed the wireframe between the floor and the right plane (detected wireframes shown as red lines). A second type of failure can occur when the image has a solid color plane. Illustrated in (b), our model segments the pure white part of the floor as part of the left/right planes. Finally, the inference might fail due to irregular planes, as shown in (c), where the buildings on the left do not form a rectangular plane. These issues could be mitigated with light user interaction, such as specifying wireframes.

\section{Conclusion}

We have presented \modelfull (\model), a framework for inferring program-like representations that model the regular texture and patterns in 2D planes and the 3D posing of these planes, all from a single image. Our model assumes a {\it box prior}, which constrains the plane segmentation of the image, and uses visual cues to guide the inference. The inferred {\program} enables 3D-aware interactive image editing, such as inpainting, extrapolation, and view synthesis.
Currently, our algorithm assumes that the full image can be partitioned into planes with regular structures. Future research may consider integrating models that can handle the presence of irregular image regions.
\section*{Broader Impact}
This paper presents an improved interactive image manipulation algorithms, which helps visual artists, photographers, and normal users who wants to perform content-aware and 3D-aware edits to their images. Moreover, our algorithm only use pixels from the input image itself during editing, which minimizes the biases coming from external sources. However, misuses of our algorithms can generate fake images that affect image forensics. 
\section*{Acknowledgements}
This work is supported by the Center for Brains, Minds and Machines (NSF STC award CCF-1231216), NSF \#1447476, ONR MURI N00014-16-1-2007, and IBM Research. Work was done while Jiajun Wu was a visiting researcher at Google Research.

{\small
\bibliographystyle{plainnat}
\bibliography{jiajun,p3i,ic}
}

\clearpage
\appendix

\begin{center}
    \LARGE \bf Supplemental Material
\end{center}

This supplementary document is organized as the following. First, in \sectapp{sec:proof}, we show the mathematical details of how to reconstruct the 3D positions and surface normal vectors of different planes based on the plane segmentation for a box's inner views (see \sectapp{sec:building} for outer views). Second, in \sectapp{sec:building}, we present the \modelfull (\model) applied to \textit{outer views} of boxes (details on \model applied to \textit{inner views} are in the main text). Next, in \sectapp{sec:detail}, we discuss the implementation details of \model. Finally, we present more qualitative results on box program induction, plane segmentation, image inpainting, and image extrapolation in \sectapp{sec:moreresults}.

\section{Plane Reconstruction from Segmentation in \textit{Inner Views}}
\label{sec:proof}
\begin{figure}[!tp]
    \centering
    \includegraphics[width=\linewidth]{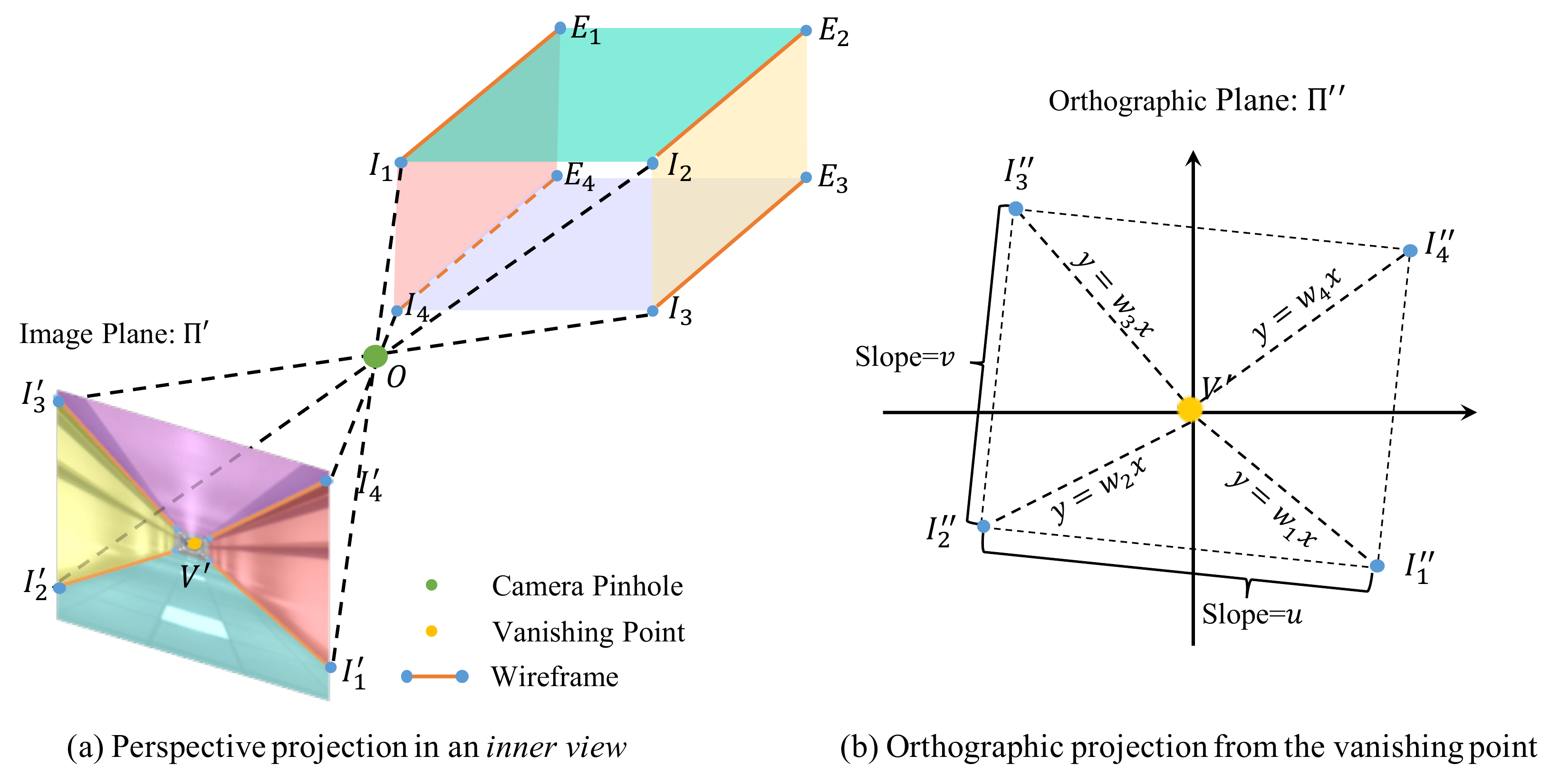}
    \caption{Illustration of (a) the perspective projection in an inner view of the box (image upside down to be consistent with the projection), and (b) the orthographic projection centered at the vanishing point.}
    \label{fig:ray2plane}
\end{figure}

In an inner view of a box, we use the plane segmentation of the input image to determine the 3D positions and surface normal vectors of four planes.
The plane segmentation is represented as four rays starting from the detected vanishing point. Our reconstruction assumes a pinhole camera model with no lens distortion, as illustrated in \fig{fig:ray2plane}a.

We start by defining the 3D coordinate system. We define the position of the camera pinhole $O$ as origin of the coordinate system. We also define the optical axis of the camera (\ie, the ray from the center of the image plane $\Pi'$ to $O$) as the $+z$ axis.

$V'$ denotes the vanishing point shown on the image plane. Four rays starting from $V'$ will intersect with the image boundary at four points $\{I'_k\mid k=1,2,3,4\}$. The line segments $\{V'I'_k\mid k=1, 2, 3, 4\}$, namely the intersection line segments, are 2D projections of the intersection lines in 3D, between four planes. We also denote these 3D intersections lines as $\{I_kE_k\mid i=1,2,3,4\}$, where $I_k$ is the corresponding 3D projection of $I'_k$, and $E_k$ is the 3D projection of an arbitrary point on the 2D line segment $V'I'_k$. Thus, all lines $\{I'_kI_k \mid k=1,2,3,4\}$ should intersect at the optical center $O$.

As can be seen in \fig{fig:ray2plane}a, the focal length (\ie the distance between the image center and the optical center $O$) is correlated with the distance between $I_k$ and $O$ (\ie, the camera-to-subject distance)\footnote{Formally, we need three planes that are perpendicular to each other to determine the focal length based on the perspective effect. However, we have only two such planes here. See Liebowitz~\etal \citep{liebowitz1999creating} for a detailed discussion.}. Moreover, the aspect ratio of the image sensor is correlated with the ratio between the sizes of four planes (\ie, the ``aspect ratio'' of the box). Thus, it is impossible to fully determine the focal length and the camera aspect ratio from this single image. Given this ambiguity, we assume the focal length to be 35mm and the aspect ratio to be 1, and then optimize for the equivalent distance between $I_k$ and $O$.

To this end, we first consider the following property of vanishing point: the line $V'O$ should be parallel with all 3D intersection lines $I_kE_k$. Thus, we perform an {\it orthographical projection} of the inner view using a new optical axis $V'O$. This leads to a new image $\Pi''$, as illustrated in \fig{fig:ray2plane}b. The line segment $OI_k$ will also be projected onto $\Pi''$ as a line segment $OI''_k$, and four intersection lines $I_kE_k$ will become four points on $\Pi''$. Determining the distance between $O$ and $I_k$ is equivalent to determining the distance between $I''_k$ and $V'$ on the new image plane $\Pi''$.

Unfortunately, one can not fully determine the distance between $I''_k$ and $V'$, but only the ratio between $V'I''_1$ and $V'I''_k, k=2,3,4$, even if we have assumed the focal length and the aspect ratio. This is because this camera-to-subject distance is also correlated with the actual size of the box. Intuitively, a box that is close and small may look identical in the image as another box that is far but big.
Thus, we will manually set the distance between $V'$ and $I''_1$ to be 1 meter. It is important to note that, although we have manually set the focal length, the aspect ratio, and the camera-to-subject distance, these value of these parameters will not affect the plane rectification results. Moreover, they also have no influence on downstream tasks such as image inpainting.

With the position of $I''_1$, we then determine the positions of $I''_k, k=2,3,4$, relative to the position of $I''_1$. We use the following {\it box prior}: four planes of the box are either perpendicular or parallel to each other. Thus, on the orthographic image plane $\Pi''$, the quadrilateral $I''_1I''_2I''_3I''_4$ should be a rectangle.

Now we use the 2D coordinates defined on the plane $\Pi''$ and centered at the vanishing point $V'$. Denote the coordinates of $I''_k$ on $\Pi''$ by $(x_k, y_k)$, the slope of $I''_1I''_2$ by $u$, and the slope of $I''_2I''_3$ by $v$. Also denote the slope of $VI''_k$ by $w_k, k=1,2,3,4$, as illustrated in \fig{fig:ray2plane}b. We have the following equation system:
\begin{equation*}
    \begin{cases}
        y_k = w_k x_k;\;k=1,2,3,4 & \text{($I''_k$ lies on the ray $VI''_k$.)}\\
        y_1-y_2 = u (x_1-x_2) & \text{(definition.)}\\
        y_2-y_3 = v (x_2-x_3) & \text{(definition.)}\\
        y_3-y_4 = u (x_3-x_4) & \text{($I''_1I''_2$ is parallel to $I''_3I''_4$.)}\\
        y_4-y_1 = v (x_4-x_1) & \text{($I''_2I''_3$ is parallel to $I''_4I''_1$.)}\\
        x_1^2 + y_1^2 = 1 & \text{(camera-to-subject distance assumption.)}\\
        u v = -1 & \text{($I''_1I''_2$ and $I''_2I''_3$ are perpendicular to each other.)}\\
    \end{cases}
\end{equation*}
This equation system allows us to solve for $u$ and $v$ unambiguously. In fact, there exists a closed-form solution to the values of $u$ and $v$, which is independent of $(x_k, y_k)$. After determining $u$ and $v$, we can further compute the positions of $I''_k$ and thus the 3D positions of $I_k,k=1,2,3,4$.

During inference, we first compute the orthographic projection $\Pi''$ based on the detected vanishing point and the focal length. Next, by fixing the position of $I''_1$, we solve for the other $I''_k, k=2,3,4$ on the orthographic image. Finally, we project the solution back to the 3D space and determine the 3D positions and surface normal vectors for individual planes.

\section{\Problemfull for \textit{Outer Views} of Boxes}
\label{sec:building}

In this section, we present the \problemfull for \textit{outer views} of boxes. The whole process is almost identical to the {\it inner view} case, except that for an outer view, we only need to consider two planes (\eg, two side planes of a building), with the other planes being either  non-visible or foreshortened severely. The full search process consists of four steps. First, we use pre-trained neural networks to detect the 2D wireframe line segments (but no vanishing points) from the image. We also filter out wireframe segments that are too short in length. Next, we generate a set of candidate plane segmentation maps by partitioning the image based on the detected wireframe segments. Then, for each segmented plane, we seek the program that best describes the regularity on each plane. Finally, we rank all candidate plane segmentations by the fitness score (defined in the main text).

\myparagraph{Step 1: Visual cue detection.}
Following the \textit{box prior}, an outer view of a box contains only two planes. Therefore, there will be only one intersection line between these two planes (\eg, two walls of a building). We use L-CNN~\citep{zhou2019end} to extract wireframes in the image. Next, we filter out wireframe segments whose length is smaller than a threshold $\delta_{1}$. The remaining wireframe segments are denoted by the set $\mathit{WF}$. Note that unlike the inner view case, we do not use vanishing point detection for outer views.

\myparagraph{Step 2: Plane segmentation.}
Next, we then extend every wireframe segment to a line. Each line will partition the input image into two parts, which we treat as the candidate plane segmentation of the image.

\myparagraph{Step 3: Plane rectification and regularity inference.}
Since we only have two planes, we cannot use the plane segmentation to fully reconstruct the positions and surface normal vectors of different planes. We run the Perspective Plane Program Induction (P3I) algorithm~\citep{Li2020Perspective} on each plane to jointly infer the surface normal of each plane and its regularity structure.

\myparagraph{Step 4: Box program ranking.}
Identical to the inner view case, we sum up the fitness score for two planes in each candidate segmentation as the overall program fitness. We use this score to rank all candidate segmentations, and pick the program with the highest fitness to describe the entire image.

\section{Implementation Details}
\label{sec:detail}
When filtering wireframes by length, we set $\delta_{1} = 0.1\times \text{min}(w, h)$, where $w, h$ are the width and height of the input image, respectively. Radius used to filter wireframes toward the detected vanishing point is $\delta_{2} = 0.01\times \text{min}(w, h)$. In the plane rectification step, we rectify the plane region to a $200\times 200$ image, on which we infer the regularity. We assume that objects repeat at least 3 times on each plane.

\section{Additional Results}
\label{sec:moreresults}
\vspace{5pt}

\myparagraph{Time complexity.}
For the corridor dataset, statistically, each image contains 1,506 wireframe combinations on average. 46 programs are evaluated on each plane. Note that \model reduces the search space significantly  based on the box prior, so that the search can be done efficiently (23$\times$ faster than without the box prior).

We also show the runtime of different algorithms on the task of plane segmentation (\tbl{tab:supp_runtime_partition}) and image inpainting (\tbl{tab:supp_runtime_inpaint}). The Image Melding~\citep{Darabi12:ImageMelding12} and \StructCompletion baselines are tested on a single machine with an Intel i7-6500U@2.5GHz CPU and 8GB RAM. All other baselines are tested on a single machine with an Intel E5-2650@2.20GHz CPU, a GeForce GTX 1080  GPU, and 8GB RAM. 

\begin{figure}[t]
    \centering
    \includegraphics[width=\linewidth]{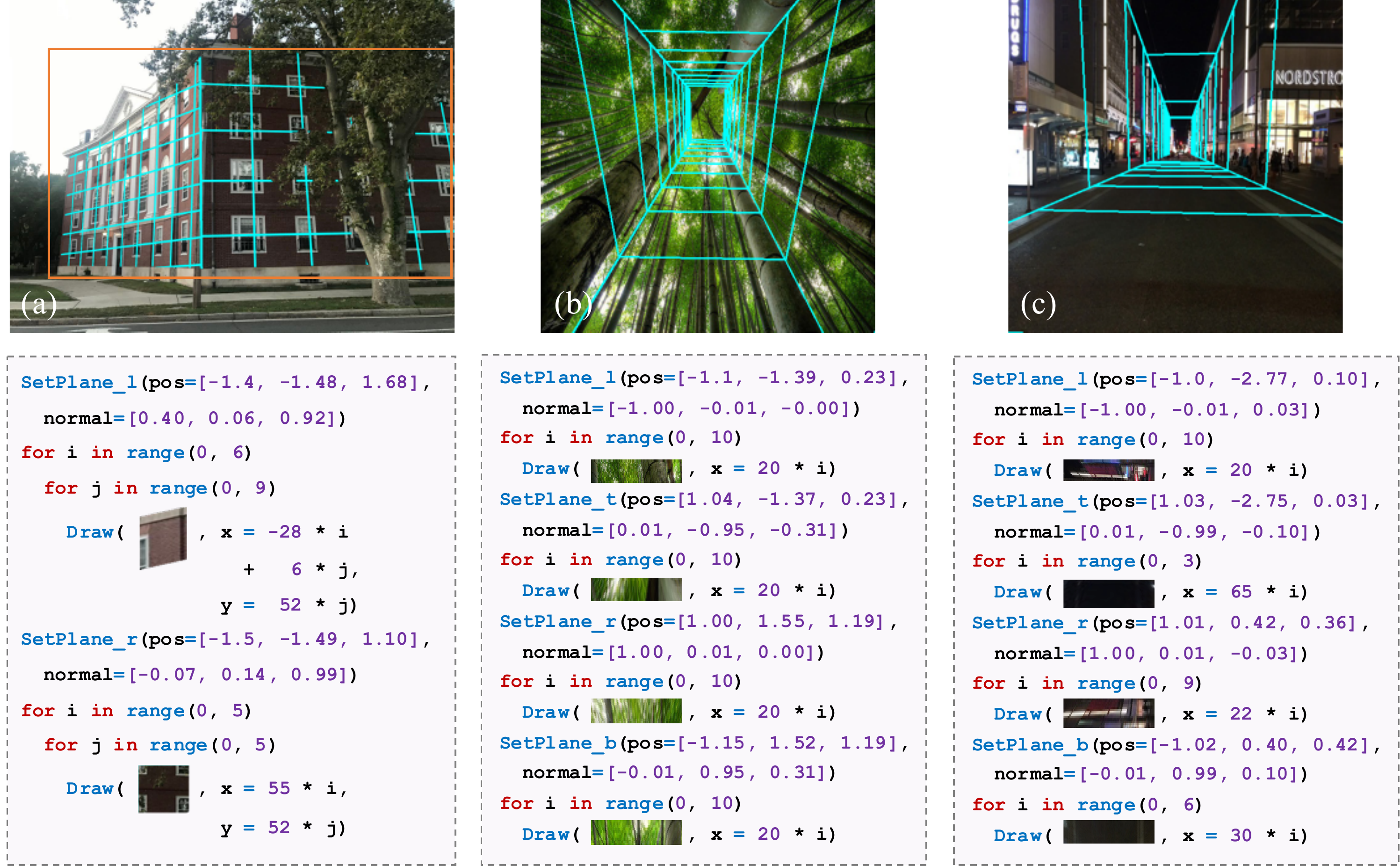}
    \caption{More image examples and the corresponding programs. We use cyan lines to visualize the lattice structure on each plane.}%
    \label{fig:supp_programs}
\end{figure}

\myparagraph{Qualitative results.}
We supplement more results on box program induction (\fig{fig:supp_programs}). Our model can be applied to less constrained images by allowing the user to specify the regular region. It also works for images where not all planes of a box exhibit regular patterns. \fig{fig:supp_programs} (i) shows example results. In (a), we run \model on a user-specified region (the orange bounding box). \model outputs a reasonable program even though the left plane is curved. We also show that our model can be applied to a broader class of images than buildings, such as the bamboo forest in (b), and the scene with irregular planes in (c).

\begin{figure}[t]
    \centering
    \includegraphics[width=\linewidth]{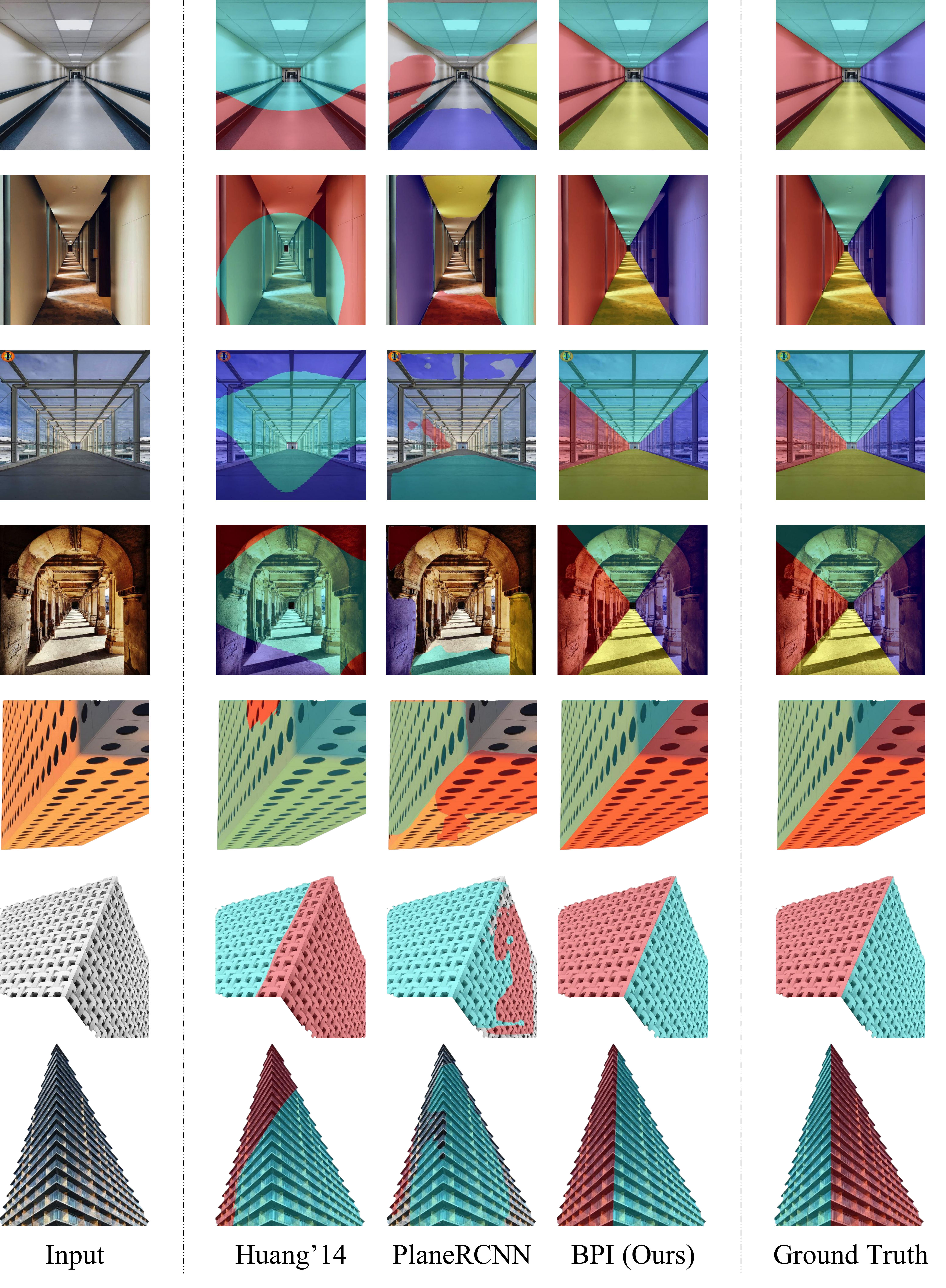}
    \caption{Visualization of the plane segmentation by different methods.}%
    \label{fig:supp_partition}
\end{figure}

\begin{figure}[t]
    \centering
    \includegraphics[width=\linewidth]{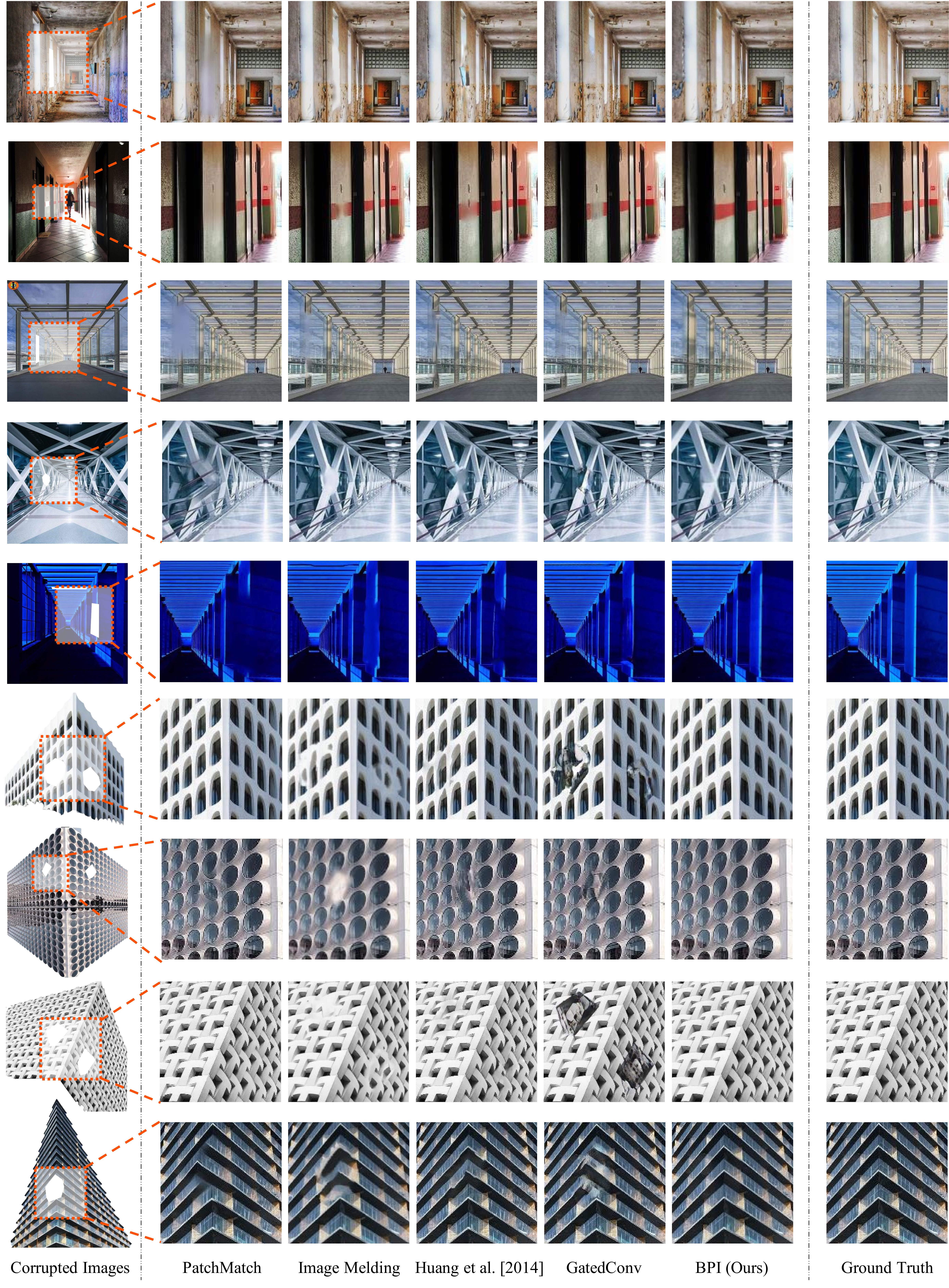}
    \caption{Qualitative results on the task of image inpainting. 
    }
    \label{fig:supp_inpaint}
\end{figure}

\begin{figure}[t]
    \centering
    \includegraphics[width=\linewidth]{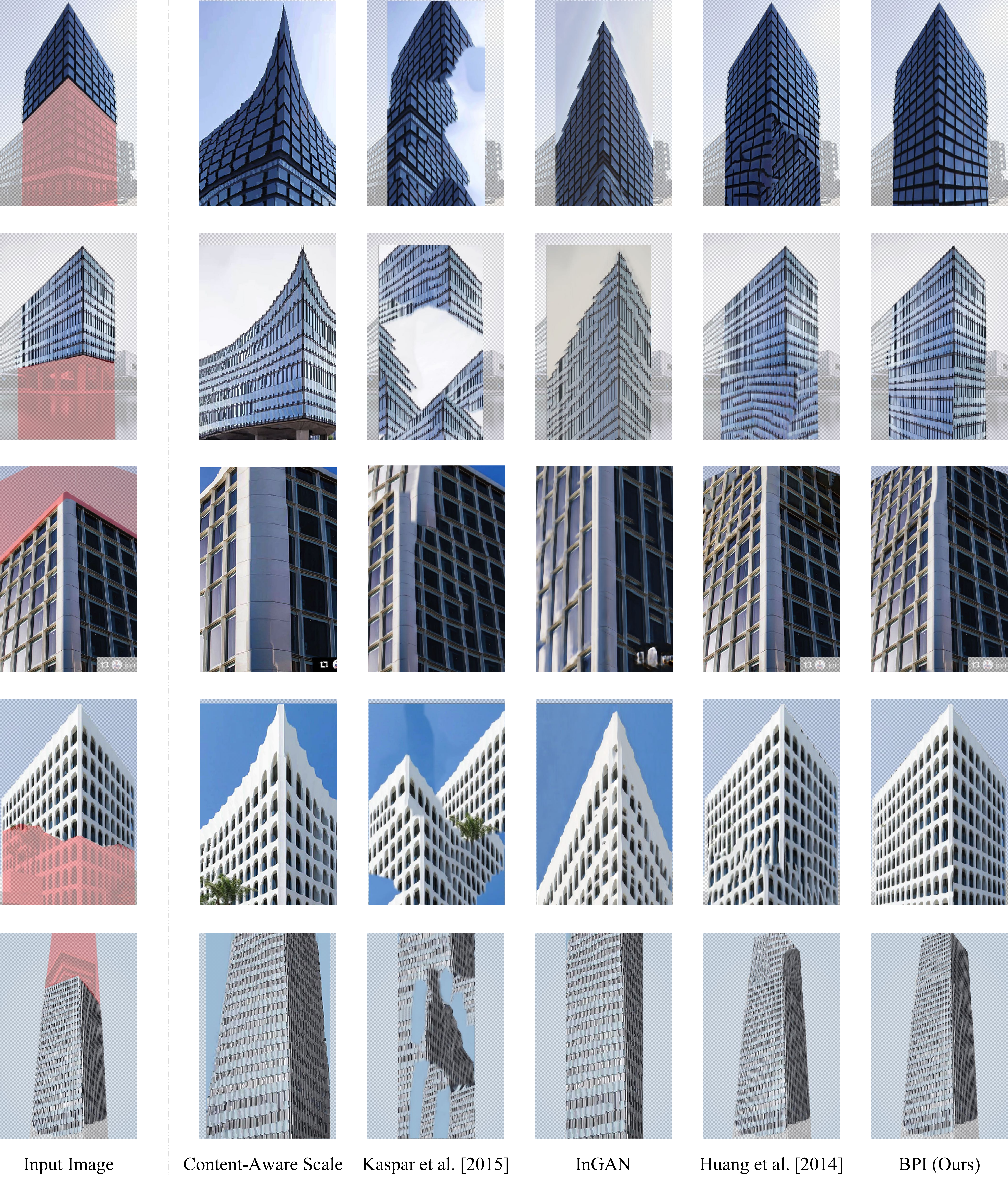}
    \caption{Qualitative results on the task of image extrapolation. 
    }
    \label{fig:supp_extrapolation}
\end{figure}

We also provide more qualitative results for plane segmentation (\fig{fig:supp_partition}), image inpainting (\fig{fig:supp_inpaint}) and image extrapolation (\fig{fig:supp_extrapolation}).

\begin{table}[t]
    \centering
    \setlength\tabcolsep{3pt}
    \begin{tabular}{lccc}
        \toprule
        &  \StructCompletion & PlaneRCNN~\citep{Liu_2019_CVPR} & \model (Ours) \\
        \midrule
        Corridor Boxes & 1.91s & 0.14s & 43.50s \\
        Building Boxes & 6.83s & 0.76s & 171.20s \\
        \arrayrulecolor{black}\bottomrule
    \end{tabular}
    \vspace{5pt}
    \caption{Runtime of different methods on the task of plane segmentation.}
    \label{tab:supp_runtime_partition}
\end{table}

\begin{table}[t]
    \centering
    \setlength{\tabcolsep}{2pt}
    \begin{tabular}{lccccc}
        \toprule
        &  PatchMatch~\citep{barnes2009patchmatch} & Image Melding~\citep{Darabi12:ImageMelding12} & \StructCompletion & Gated Conv~\citep{yu2019free} & \model (Ours) \\
        \midrule
        Corridor Boxes & 61.6s & 1275.1s & 18.9s & 1.9s & 20.1s \\
        Building Boxes & 122.1s & 808.3s & 64.4s & 4.9s & 41.4s \\
        \arrayrulecolor{black}\bottomrule
    \end{tabular}
    \vspace{5pt}
    \caption{Runtime of different methods on the task of image inpainting.}
    \label{tab:supp_runtime_inpaint}
\end{table} 
\newpage


\end{document}